%% file: AnonymousSubmission2027.tex
\documentclass[letterpaper]{article} 
\usepackage[preprint]{aaai2027}  
\usepackage[hyphens]{url}  
\usepackage{graphicx} 
\usepackage{natbib}  
\usepackage{caption} 
\usepackage{algorithm}
\usepackage{algorithmic}

\usepackage{newfloat}
\usepackage{listings}
\DeclareCaptionStyle{ruled}{labelfont=normalfont,labelsep=colon,strut=off} 
\floatstyle{ruled}
\newfloat{listing}{tb}{lst}{}
\floatname{listing}{Listing}

\usepackage{booktabs}

\usepackage{amssymb}
\usepackage{amsmath}
\usepackage{multirow}
\usepackage{booktabs}
\usepackage{xcolor}

\definecolor{revblue}{RGB}{0,70,180}

\title{One Offset, Two Directions: Baseline Adjustment via Subgroup Offsets in GRPO}
\author{
    Jiaying Zhou\textsuperscript{\rm 1}\thanks{Work done during an internship at JD.com.},
    Jiachen Liu\textsuperscript{\rm 2},
    Qingchao Chen\textsuperscript{\rm 1}
}
\affiliations{
    \textsuperscript{\rm 1}Peking University \quad
    \textsuperscript{\rm 2}JD.com
}

\begin{document}

\maketitle

\input{sections/0_abstract}
\input{sections/1_intro}
\input{sections/2_related}
\input{sections/3_method}

\input{sections/4_experiment}
\input{sections/5_conclusion}





\nocite{*}
\bibliography{aaai2027}


\clearpage

\end{document}

%% file: sections/0_abstract.tex
\begin{abstract}
Group Relative Policy Optimization (GRPO) estimates advantages against the within-prompt mean reward, a baseline shared across the group. 
In practice, rollouts within a prompt are not always homogeneous, differing in their solution strategies and reasoning trajectories, along which they may group into subgroups. 
We further find that these heterogeneous subgroups may exhibit distinct reward distributions. 
The global baseline, however, collapses them onto a single credit axis, 
where the credit reflects neither a rollout's standing within its subgroup nor the subgroup's standing relative to the rest. 
We therefore study how to adjust the baseline when the rollouts of a prompt fall into heterogeneous subgroups. 
We define the \emph{subgroup offset} as the deviation of a subgroup mean from the global mean, 
and shift the baseline along it by a single signed coefficient $\lambda$.
The sign of $\lambda$ yields two opposing directions: 
\emph{Alignment} ($\lambda{>}0$) adds the offset to the baseline, drawing credit toward a rollout's relative position within its subgroup; 
\emph{Contrast} ($\lambda{<}0$) subtracts it, drawing a rollout's credit toward the relative quality of the subgroup it belongs to. 
The adjustment acts on the baseline alone, leaving the reward signal and the optimization objective unchanged. 
In this work, we induce subgroups from strategies, grouping rollouts that share a strategy into the same subgroup. 
We evaluate both directions on Qwen3-8B-Base and Olmo3-7B-Base across four competition-level math benchmarks, 
against a $\lambda{=}0$ anchor that shares the same strategy-conditioned sampling and differs only in the baseline. 
Against this anchor each direction yields modest but consistent gains in multi-sample coverage, with Contrast the stronger of the two.

\end{abstract}

%% file: sections/1_intro.tex
\section{Introduction}
\label{sec:intro}

Reinforcement Learning with Verifiable Rewards (RLVR) has become the dominant paradigm for improving the reasoning ability of Large Language Models (LLMs)~\cite{lambert2024tulu,guo2025deepseek,yang2025qwen3}.
Among RLVR algorithms, Group Relative Policy Optimization (GRPO)~\cite{shao2024deepseekmath} is widely adopted for its simplicity and stability~\cite{yu2026dapo,hu2026open}.
Instead of training an explicit critic, which is costly and hard to fit for long reasoning traces, it samples a group of rollouts for each prompt and estimates each rollout's advantage against their mean reward, which serves as the group baseline.
As a result, equally-rewarded rollouts receive the same advantage. 

\begin{figure*}[t]
\centering 
\includegraphics[width=\linewidth]{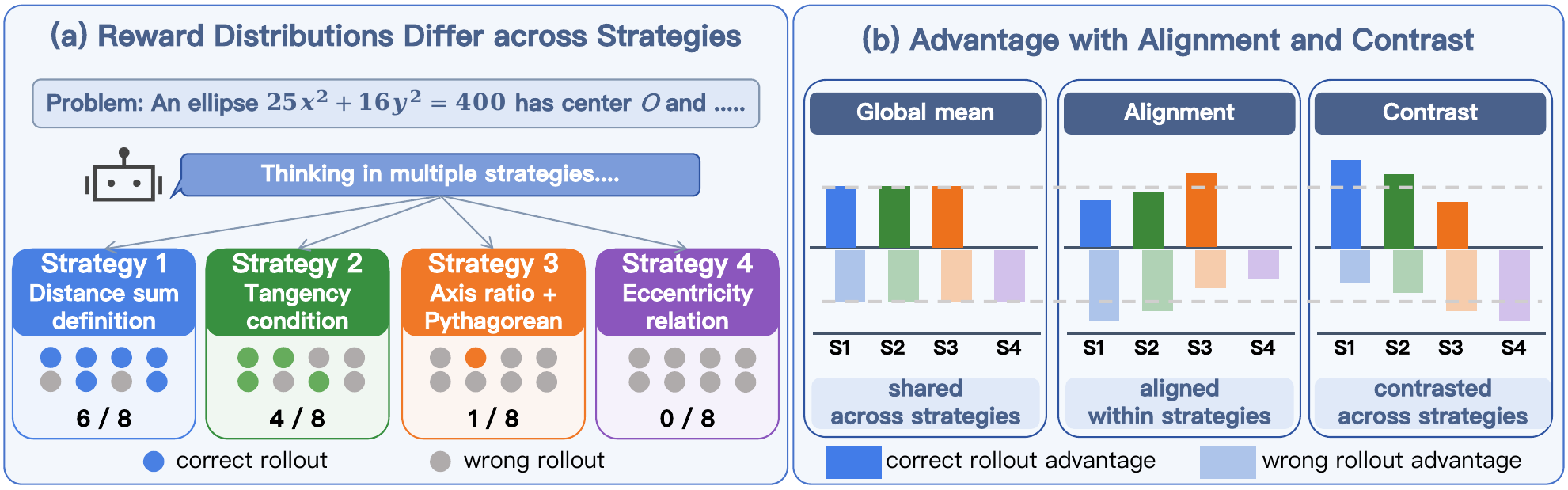}
\caption{(a) Different strategies may lead to markedly different reward distributions. For the same problem, success rates across the four strategies range from $0.75$ down to $0$ (strategies and the corresponding rollouts are sampled from Qwen3-8B-Base at an intermediate checkpoint). (b) Effect of the two adjustment directions on the advantage. Relative to the global mean, which shares one baseline across all strategies, \textit{Alignment} credits each rollout within the same strategy, while \textit{Contrast} amplifies the differences between strategies. Bar heights indicate the relative magnitude of the advantage only, not exact values. }
\label{fig:teaser}
\end{figure*}

The rollouts in a group, however, are not always homogeneous. 
They may differ in the solution strategies they adopt, the reasoning trajectories they follow, and the intermediate steps they take, and such differences carry through to different outcomes and, ultimately, different reward distributions~\cite{zhu2025stratified, dou2025plan}. 
In Figure~\ref{fig:teaser}(a), we group the rollouts of a single problem by their strategies. 
The resulting subgroups vary widely in reward distribution: under some strategies most rollouts reach the correct answer as with \textit{strategy 1}, whereas under others success is rare. 
The global baseline is blind to this structure. 
Averaging over the entire group, it places every rollout on a single axis, where credit reflects only the deviation from the group mean and disregards both a rollout's standing within its strategy and that strategy's standing among all strategies.

We therefore ask how the baseline should respond when the rollouts of a group fall into subgroups with different reward distributions. 
We define the \emph{subgroup offset} as the deviation of a subgroup's mean reward from the global mean, and shift the baseline along it by a single signed coefficient $\lambda$, whose sign determines the direction of the adjustment. 
For $\lambda{>}0$, the offset is added to the baseline. We show that the resulting advantage is a convex combination of the standard global advantage and one measured against the subgroup mean. We call this direction \textit{Alignment}, as a rollout's credit is aligned toward its subgroup. 
For $\lambda{<}0$, the offset is subtracted from the baseline. We show that the resulting advantage augments the standard global advantage with a term proportional to the subgroup offset, which serves as a proxy for the subgroup's overall quality. We call this direction \textit{Contrast}, as a rollout's credit is drawn toward the contrast between subgroups. 
Both directions act only on the baseline, with reward distribution unchanged, with $\lambda{=}0$ recovering the original baseline in standard GRPO. Figure~\ref{fig:teaser}(b) shows how the two directions move the advantages in opposite ways. 

In this work, following PTA-GRPO~\cite{dou2025plan}, we induce subgroups by conditioning each rollout on a presampled strategy, so that rollouts sharing a strategy form one subgroup.
We evaluate both directions on Qwen3-8B-Base~\cite{yang2025qwen3} and Olmo3-7B-Base~\cite{olmo2025olmo} across four competition-level math benchmarks.
Experiments show that both \textit{Alignment} and \textit{Contrast} improve modestly over the $\lambda{=}0$ baseline, with \textit{Contrast} slightly outperforming \textit{Alignment}.

Our contributions are as follows:
\begin{itemize}
\item We identify the subgroup offset under GRPO's global baseline, and show how it enters the advantage from a subgroup perspective.
\item We propose a bidirectional baseline adjustment that unifies the alignment and contrast directions under a single signed coefficient, while keeping the reward distribution and the optimization objective unchanged.
\item On Qwen3-8B-Base and Olmo3-7B-Base, we show that both directions yield modest but consistent gains over the $\lambda{=}0$ anchor, concentrated in multi-sample coverage, and verify through ablation that these gains rely on the genuine correspondence between subgroups and their offsets.
\end{itemize}

%% file: sections/2_related.tex
\section{Related Work}

\paragraph{Reinforcement learning with verifiable rewards.} 
Reinforcement learning with verifiable rewards~(RLVR) has become a key technique for improving the reasoning ability of large language models~\cite{lambert2024tulu,guo2025deepseek,yang2025qwen3}. 
Rather than relying on a learned reward model, it derives the reward from deterministic verifiers such as answer checking or program execution~\cite{guo2025deepseek,hu2026open,wang2025ragen}. 
This paradigm is most effective in domains where correctness is cheaply and reliably checkable, such as mathematics and code, on which recent large reasoning models are largely built. 
Among RLVR algorithms, GRPO~\cite{shao2024deepseekmath} is widely adopted for dispensing with the critic: rather than fitting a separate value network, it samples a group of rollouts per prompt and estimates advantages relative to the group's mean reward. 
Building on this, the subsequent DAPO~\cite{yu2026dapo} stabilizes training at scale with decoupled clipping and dynamic sampling, and has become a common recipe for reasoning systems. 
A growing body of work further refines the GRPO objective to improve exploration~\cite{cui2025entropy,cheng2026reasoning,wu2026qae}, training stability~\cite{yu2026dapo}, and sample efficiency~\cite{liu2025drgrpo,chen2025pass}.

\paragraph{Group-relative advantage estimation.} 
The wide adoption of GRPO has prompted a range of refinements to its advantage estimation, whose single shared mean baseline can render credit sparse, noisy, biased, or prone to advantage collapse~\cite{shao2024deepseekmath,le2025no}. 
These efforts largely target its three constituents: the outcome reward, the group baseline, and the normalization. 
On the reward, one line reshapes or augments the scalar reward signal to densify otherwise sparse credit~\cite{niu2026rlrr,he2026vane}. 
On the baseline, robust statistics such as the median or a group quantile replace the mean, or a tunable baseline reweights positive and negative rewards~\cite{kim2026mc,wu2026qae,arnal2026asymmetric}. 
On the normalization, others debias for length or renormalize within homogeneous strata~\cite{liu2025drgrpo,zhu2025stratified,he2026avspo}. 
Our method also acts on the baseline, reading the heterogeneous subgroups within a group as the stratified line does, but differing in the operator rather than the axis: a single signed coefficient scales the subgroup offset into the baseline, tracing a continuous bidirectional family in which $\lambda{=}0$ recovers GRPO and $\lambda{=}1$ reduces to within-subgroup centering~\cite{zhu2025stratified}. 
Prior estimators fix a single point of this family, and none covers its negative half.

\paragraph{Subgroup structure within rollout groups.} 
Recent work has looked beyond the flat rollout group and identified finer structure within it, differing in how subgroups are defined: by tool-use or execution structure~\cite{zhu2025stratified,wang2026implicit}, by presampled plans that rollouts share~\cite{dou2025plan}, by branches of a search tree~\cite{tran2025tempo,xing2025lookahead}, or by repeated states, segments, or graph structure within trajectories~\cite{feng2026gigpo,guo2026segment,cheng2026beyond}. 
We follow this view, inducing subgroups from presampled strategies. Crucially, the strategy is never optimized or fed back into the policy; it serves purely as a grouping proxy that partitions rollouts into subgroups, on which our baseline adjustment then operates.                                                          

%% file: sections/3_method.tex
\section{Method}

GRPO measures each rollout's advantage relative to the group mean, a global baseline under which equally-rewarded rollouts receive equal advantage. 
When the rollouts fall into heterogeneous subgroups, credit assigned along this single axis reflects neither a rollout's position within its subgroup nor the quality of that subgroup.  
We therefore study how to adjust the baseline when a group contains heterogeneous subgroups. 
Specifically, we first construct heterogeneous subgroups by sampling rollouts conditioned on presampled strategies; 
we then define the subgroup offset and adjust the baseline by it; 
and we trace the effect of the baseline adjustment on the advantage.

\subsection{Preliminaries}
\label{ssec:prelim}

\paragraph{Problem formulation and notation.} 
Let $x\sim\mathcal{D}$ be a prompt from distribution $\mathcal{D}$, and let $\pi_\theta$ denote the policy, a language model with parameters $\theta$. 
A verifiable reward $R(x,y)\in\mathbb{R}$, typically binary under RLVR, scores each rollout $y$, 
and the objective is to maximize the expected reward $\mathbb{E}_{x\sim\mathcal{D},\,y\sim\pi_\theta(\cdot\mid x)}[R(x,y)]$. 
For each prompt, the old policy $\pi_{\theta_{\text{old}}}$ samples a group of $N$ rollouts $\{y_i\}_{i=1}^{N}$ with rewards $\{r_i\}_{i=1}^{N}$, 
from which advantages are estimated to update $\pi_\theta$.

\paragraph{Group-relative advantage.} 
Rather than learning a separate value function, GRPO estimates each rollout's advantage by standardizing the reward within its group: 

\begin{equation}
A_i^{\mathrm{GRPO}}=\frac{r_i-\mu_g}{\sigma_g},
\label{eq:grpo-adv}
\end{equation}
where $\mu_g$ and $\sigma_g$ are the mean and standard deviation of the rewards $\{r_i\}_{i=1}^{N}$.
Every rollout is centered on the same $\mu_g$, so equally-rewarded rollouts receive equal advantage regardless of how they differ. 
When all $N$ rollouts of a group share one reward, $\sigma_g{=}0$ and we set every advantage in that group to zero, so the group contributes no gradient; the same convention carries to Eqs.~(\ref{eq:adv-mix}) and~(\ref{eq:adv-shift}).

This advantage then weights a clipped policy-gradient objective:
\begin{equation}
\mathcal{J}(\theta)=\mathbb{E}\big[\min(\rho_i A_i,\ \mathrm{clip}(\rho_i,1-\epsilon_{\mathrm{low}},1+\epsilon_{\mathrm{high}})\,A_i)\big],
\label{eq:grpo-obj}
\end{equation}
where $\rho_i$ is the importance ratio between $\pi_\theta$ and $\pi_{\theta_{\text{old}}}$, $A_i$ is the advantage from Eq.~(\ref{eq:grpo-adv}), and the clipping constrains the size of each update.

\subsection{Subgroup Construction}
\label{ssec:subgroup}
Adjusting the baseline for heterogeneous subgroups presupposes a partition of the rollouts into subgroups. 
An informative partition should keep rollouts within a subgroup relatively homogeneous while keeping subgroups distinct from one another, so that each subgroup offset is meaningful. 

Following PTA-GRPO~\cite{dou2025plan}, we obtain such a partition at low cost by conditioning rollouts on presampled strategies, without any auxiliary model. 
For each prompt $x$, we sample a pool of candidate strategies and select $K$ of them to encourage diversity across subgroups: 
\begin{equation}
s_k \sim \pi_\theta(\cdot \mid x), \quad k=1,\dots,K,
\end{equation}
and then sample $Z$ rollouts conditioned on each strategy: 
\begin{equation}
y_{k,z} \sim \pi_\theta(\cdot \mid x, s_k), \quad z=1,\dots,Z, 
\end{equation}
giving $K$ subgroups and $N=KZ$ rollouts in total. 
Each subgroup $k$ has mean reward $\mu_k=\frac{1}{Z}\sum_{z=1}^{Z} r_{k,z}$, and we write $\mu_{c(i)}$ for the mean of the subgroup containing rollout $i$. 
Unlike PTA-GRPO, we neither reward nor optimize the strategy; 
it serves only to induce subgroups.

\subsection{Subgroup Offset and Baseline Adjustment}
\label{ssec:adjust}
The group-mean baseline centers every rollout on a single value $\mu_g$, treating the group as homogeneous. 
When the rollouts instead fall into heterogeneous subgroups of differing reward distributions, this single baseline carries no information about a rollout's subgroup. 
We capture this with the \emph{subgroup offset}, the deviation of a rollout's subgroup mean from the global mean: 
\begin{equation}
\delta_i=\mu_{c(i)}-\mu_g, 
\label{eq:offset}
\end{equation} 
shared by all rollouts in the same subgroup and summing to zero over the group. 

We then use this offset to adjust the baseline for each subgroup, scaling it by a single signed coefficient $\lambda$:
\begin{equation} 
\tilde b_i(\lambda)=\mu_g+\lambda\,\delta_i. 
\label{eq:baseline}
\end{equation} 
The sign of $\lambda$ defines two directions of relocating each subgroup's reference. 
For $\lambda>0$, each baseline moves in the direction of its own offset: subgroups above the global mean are assigned a higher reference and those below a lower one, so that $\tilde b_i$ shifts toward the subgroup mean $\mu_{c(i)}$ and reaches it exactly at $\lambda=1$. 
For $\lambda<0$, each baseline moves against its offset: above-average subgroups are assigned a lower reference and below-average ones a higher reference, so that $\tilde b_i$ is pushed away from its own subgroup mean and past the global value. 
In both directions the spread of references across subgroups widens with $|\lambda|$; the sign fixes whether this spread preserves the ordering of subgroup means or inverts it. 
Setting $\lambda=0$ collapses every reference back onto the shared global baseline $\mu_g$.

\begin{table*}[t]
\centering
\begin{tabular}{lcccccccc}
\toprule
\multirow{2}{*}{Method} & \multicolumn{2}{c}{AIME 2024} & \multicolumn{2}{c}{AIME 2025} & \multicolumn{2}{c}{AIME 2026} & \multicolumn{2}{c}{HMMT 2025} \\
\cmidrule(lr){2-3}\cmidrule(lr){4-5}\cmidrule(lr){6-7}\cmidrule(lr){8-9} &  pass@1 & pass@16 & pass@1 & pass@16 & pass@1 & pass@16 & pass@1 & pass@16 \\
\midrule
\textit{Qwen3-8B-Base} &  &  &  &  &  &  &  & \\
\textit{GRPO} & \textit{31.6} & \textit{66.8} & \textit{24.0} & \textit{50.3} & \textit{25.5} & \textit{58.3} & \textit{12.8} & \textit{32.5} \\
Strategy-GRPO & 30.2 & 65.6 & 23.0 & 47.9 & 23.5 & 53.8 & \underline{12.6} & 32.5 \\
Alignment & \underline{31.1} $\uparrow$ & \underline{66.4} $\uparrow$ & \textbf{24.0} $\uparrow$ & \underline{49.8} $\uparrow$ & \underline{24.4} $\uparrow$ & \underline{56.0} $\uparrow$ & 12.1 $\downarrow$ & \underline{32.7} $\uparrow$ \\
Contrast & \textbf{31.8} $\uparrow$ & \textbf{69.1} $\uparrow$ & \underline{23.8} $\uparrow$ & \textbf{50.6} $\uparrow$ & \textbf{24.6} $\uparrow$ & \textbf{58.8} $\uparrow$ & \textbf{13.0} $\uparrow$ & \textbf{36.3} $\uparrow$ \\
\midrule
\textit{Olmo3-7B-Base} &   &  &   &   &   &   &   & \\
\textit{GRPO} & \textit{42.3} & \textit{72.9} & \textit{32.4} & \textit{57.6} & \textit{33.5} & \textit{62.3} & \textit{16.4} & \textit{37.1} \\
Strategy-GRPO & 38.8 & 69.8 & 30.4 & 54.6 & \underline{30.3} & 62.2 & \textbf{16.3} & 34.7 \\
Alignment & \underline{38.8} & \underline{71.4} $\uparrow$ & \underline{31.2} $\uparrow$ & \textbf{56.4} $\uparrow$ & 29.6 $\downarrow$ & \underline{62.9} $\uparrow$ & \underline{15.4} $\downarrow$ & \underline{36.1} $\uparrow$ \\
Contrast & \textbf{40.1} $\uparrow$ & \textbf{72.9} $\uparrow$ & \textbf{31.5} $\uparrow$ & \underline{56.3} $\uparrow$ & \textbf{30.5} $\uparrow$ & \textbf{65.2} $\uparrow$ & 14.6 $\downarrow$ & \textbf{37.3} $\uparrow$ \\
\bottomrule
\end{tabular}
\caption{Main results on four competition-level math benchmarks (pass@1 / pass@16). Arrows denote the change relative to Strategy-GRPO ($\lambda{=}0$), the anchor shared by both directions, which is the comparison that isolates the baseline adjustment. The best and second-best results among the strategy-based methods are shown in bold and underlined, respectively; the GRPO row is shown in italics for reference.}
\label{tab:main}
\end{table*}

\subsection{From Adjustment to Credit: \textit{Alignment} and \textit{Contrast}}
\label{ssec:direction}

The adjustment enters only through the baseline, but its effect is realized in the advantage. 
We now derive this effect, tracing how this adjustment reshapes each rollout's advantage. 
With the offset carried through, the normalized advantage can be expanded as
\begin{align} 
A_i(\lambda)
&=\frac{r_i-\tilde b_i(\lambda)}{\sigma_g} 
=\frac{r_i-\mu_g-\lambda\,\delta_i}{\sigma_g} \notag\\  
&=\frac{(1-\lambda)\,(r_i-\mu_g)+\lambda\,(r_i-\mu_{c(i)})}{\sigma_g} \notag\\ 
&=(1-\lambda)\,A_i^{\mathrm{glob}}+\lambda\,A_i^{\mathrm{loc}},
\label{eq:adv-mix} 
\end{align}
where
\begin{equation}
A_i^{\mathrm{glob}}=\frac{r_i-\mu_g}{\sigma_g},
\qquad
A_i^{\mathrm{loc}}=\frac{r_i-\mu_{c(i)}}{\sigma_g},
\label{eq:axes}
\end{equation}
both normalized by the global standard deviation $\sigma_g$:  $A_i^{\mathrm{glob}}$ is the standard GRPO advantage, while $A_i^{\mathrm{loc}}$ centers each rollout on its subgroup mean $\mu_{c(i)}$. 

For $0<\lambda<1$, Eq.~\eqref{eq:adv-mix} is a convex combination of $A_i^{\mathrm{glob}}$ and $A_i^{\mathrm{loc}}$, so that each rollout's advantage combines a within-group and a within-subgroup component.
Larger $\lambda$ shifts weight onto the within-subgroup component, so credit increasingly reflects a rollout's position within its subgroup, reducing to $A_i^{\mathrm{loc}}$ alone at $\lambda{=}1$. 
We call this direction \textit{Alignment}, as credit is aligned with the subgroup a rollout belongs to.
Such alignment is reasonable because whether a rollout succeeds depends partly on the strategy it follows;
crediting it within its subgroup factors out this shared component, isolating the rollout's own contribution.

The advantage admits a second equivalent form, isolating the offset as a separate term: 
\begin{align}
A_i(\lambda)
&=\frac{r_i-\tilde b_i(\lambda)}{\sigma_g} 
=\frac{r_i-\mu_g-\lambda\,\delta_i}{\sigma_g} \notag\\ 
&=A_i^{\mathrm{glob}}-\lambda\,\frac{\delta_i}{\sigma_g}.
\label{eq:adv-shift} 
\end{align}
$\delta_i/\sigma_g$ is the normalized subgroup offset, so that the second term rescales credit by how a subgroup compares with the group. 

For $\lambda<0$, the offset term $-\lambda\,\delta_i/\sigma_g$ in Eq.~\eqref{eq:adv-shift} shares the sign of $\delta_i$, raising the advantages of rollouts in above-average subgroups and lowering those in below-average ones, and thereby amplifying the disparity between subgroups. 
We call this direction \textit{Contrast}, as credit reflects a subgroup's contrast with the group. 
Such contrast is reasonable: a subgroup's mean reward reflects the quality of its strategy~\cite{dou2025plan}, and we take the offset as a proxy for a subgroup's relative quality. Contrast then concentrates positive credit on rollouts from the more successful subgroups, steering the policy toward the strategies more likely to succeed. 
Both directions act on the baseline alone, not on the reward: as with GRPO's group mean, the baseline is formed from the in-group rewards, so the estimator keeps the same self-normalized form as GRPO rather than introducing a new source of bias.

Since $\delta_i$ is computed from the $Z$ rewards of one subgroup, it carries a standard error of up to $0.18$ at $Z{=}8$, so $|\lambda|$ governs how much estimation noise accompanies the subgroup structure it admits.

%% file: sections/4_experiment.tex
\begin{table*}[t]
\centering
\begin{tabular}{lcccccccc}
\toprule
\multirow{2}{*}{$\lambda$} & \multicolumn{2}{c}{AIME 2024} & \multicolumn{2}{c}{AIME 2025} & \multicolumn{2}{c}{AIME 2026} & \multicolumn{2}{c}{HMMT 2025} \\
\cmidrule(lr){2-3}\cmidrule(lr){4-5}\cmidrule(lr){6-7}\cmidrule(lr){8-9}
& pass@1 & pass@16 & pass@1 & pass@16 & pass@1 & pass@16 & pass@1 & pass@16 \\ 
\midrule
$0$ (Strategy-GRPO) & 30.2 & 65.6 & 23.0 & 47.9 & 23.5 & 53.8 & 12.6 & 32.5 \\
\midrule 
\multicolumn{9}{l}{\textit{Alignment} ($\lambda>0$)} \\
$+0.3$ & \textbf{31.1} & \textbf{66.4} & \textbf{24.0} & \textbf{49.8} & \underline{24.4} & \textbf{56.0} & \textbf{12.1} & \textbf{32.7} \\
$+0.5$ & 30.3 & 64.3 & 23.6 & 46.8 & 23.6 & 51.5 & 11.6 & 29.0 \\ 
$+0.7$ & \underline{30.9} & \underline{65.8} & \underline{23.9} & \underline{48.4} & \textbf{24.8} & \underline{54.1} & \underline{12.0} & \underline{29.2} \\ 
\midrule
\multicolumn{9}{l}{\textit{Contrast} ($\lambda<0$)} \\
$-0.3$ & \textbf{31.8} & \textbf{69.1} & \textbf{23.8} & \textbf{50.6} & \textbf{24.6} & \textbf{58.8} & \textbf{13.0} & \textbf{36.3} \\
$-0.5$ & \underline{31.1} & 65.0 & \underline{22.8} & \underline{48.2} & \underline{24.4} & \underline{55.0} & \underline{11.7} & \underline{31.9} \\
$-0.7$ & 30.1 & \underline{65.2} & 22.4 & 47.3 & 23.6 & 49.4 & 9.8 & 26.0 \\
\bottomrule 
\end{tabular} 
\caption{Effect of the coefficient $\lambda$ on Qwen3-8B-Base (pass@1 / pass@16). We vary $\lambda$ on both sides of the $\lambda{=}0$ anchor (Strategy-GRPO) to probe how the two directions respond to the strength of the baseline adjustment. For each column, the best result is in \textbf{bold} and the second best is \underline{underlined}.} 
\label{tab:lambda} 
\end{table*}

\section{Experiments}

We evaluate the two directions of our baseline adjustment on Qwen3-8B-Base~\cite{yang2025qwen3} and Olmo3-7B-Base~\cite{olmo2025olmo} across four competition-level math benchmarks. 
Compared with the $\lambda{=}0$ baseline, \textit{Alignment} and \textit{Contrast} each yield modest but consistent gains across models, with \textit{Contrast} slightly stronger. 
We then ablate the coefficient $\lambda$ and the correspondence between subgroups and their offsets, finding that the two directions respond differently to $\lambda$ and that the gains no longer hold once the correspondence is broken. 
Finally, we analyze the training dynamics of the two directions, and discuss our findings.

\subsection{Experimental Setup}

\paragraph{Models and training.}
We train on Qwen3-8B-Base~\cite{yang2025qwen3} and Olmo3-7B-Base~\cite{olmo2025olmo} with DAPO-Math-17K~\cite{yu2026dapo}, a set of about 17K math problems with verifiable answers. 
Training follows the DAPO~\cite{yu2026dapo} recipe with dynamic sampling disabled, using a global batch size of 256 and a mini-batch size of 32. 
For each prompt we first sample a pool of candidate strategies and select $K{=}4$ of them, then sample $Z{=}8$ rollouts conditioned on each strategy, giving $N{=}32$ rollouts in total; 
both strategies and rollouts are sampled at temperature $1.0$. 
Unless otherwise noted, \textit{Alignment} uses $\lambda{=}0.3$ and \textit{Contrast} uses $\lambda{=}-0.3$. 
All runs share the same training configuration. 
Full implementation details, including the clipping thresholds, learning rate, prompts and strategy-selection procedure, among others, are provided in the appendix due to space limits.

\paragraph{Evaluation benchmarks.} 
We evaluate on four competition-level math-reasoning benchmarks: AIME24, AIME25, AIME26, and HMMT25. 
Evaluation is single-stage, without the strategy-conditioning used in training. 
As these benchmarks contain few problems, 
we replicate each problem 32 times to obtain stable estimates; for each replica we sample 16 rollouts at temperature $0.7$. 
We report pass@1, the mean accuracy over the 16 rollouts, and pass@16, the fraction of problems solved by at least one of them, averaging both over the replicas.

\paragraph{Compared methods.} 
All methods share the training recipe above and differ only in whether rollouts are grouped into subgroups and how the baseline is adjusted: 
\begin{itemize}
    \item \textbf{GRPO} samples rollouts directly, without strategies, and centers each on the group mean.
    \item \textbf{Strategy-GRPO} groups rollouts by strategy but still centers each on the group mean ($\lambda{=}0$), serving as the anchor for both directions. 
    \item \textbf{\textit{Alignment}} ($\lambda{>}0$) adds the offset to the baseline, drawing each rollout's credit toward its own subgroup.  
    \item \textbf{\textit{Contrast}} ($\lambda{<}0$) subtracts the offset instead, pulling credit apart between subgroups. 
\end{itemize}

\subsection{Main Results}
Table~\ref{tab:main} summarizes results on Qwen3-8B-Base and Olmo3-7B-Base across the four benchmarks. 
Both \textit{Alignment} and \textit{Contrast} improve over Strategy-GRPO, the $\lambda{=}0$ anchor, with clear gains in pass@16 and smaller gains in pass@1. 
On Qwen3-8B-Base, \textit{Contrast} is the stronger direction, improving pass@16 by $3.8$ points and pass@1 by $1.0$ on average; 
\textit{Alignment} also helps, though more modestly ($+1.3$ and $+0.6$). 
On Olmo3-7B-Base the pattern repeats: \textit{Contrast} improves pass@16 by $2.6$ points and pass@1 by $0.2$, 
while \textit{Alignment} improves pass@16 by $1.4$ and slightly lowers pass@1 ($-0.2$). 
Across both models, Contrast is consistently the stronger direction, and the gains fall mainly on pass@16 rather than pass@1. 

We also note that Strategy-GRPO ($\lambda{=}0$) trails plain GRPO on both models, more so on Olmo3-7B-Base.
This is expected, since conditioning each prompt on a sampled strategy shifts the training distribution, whereas evaluation is single-stage and queries the model directly, leaving a mismatch between training and test prompts.
Grouping by strategy is not intended to raise accuracy on its own; it is the mechanism that yields subgroups and makes the baseline adjustment possible.

The two comparisons therefore answer different questions. Against the $\lambda{=}0$ anchor, which holds the sampling fixed and varies only the baseline, the adjustment is a modest but consistent win, and this is the comparison that isolates our contribution. Against plain GRPO, Contrast closes the pass@16 gap and ends ahead on all four benchmarks of Qwen3-8B-Base, at pass@1 that is level with GRPO; on Olmo3-7B-Base it recovers pass@16 only on average and remains behind on pass@1.

\begin{figure*}[t] 
\centering
\includegraphics[width=\linewidth]{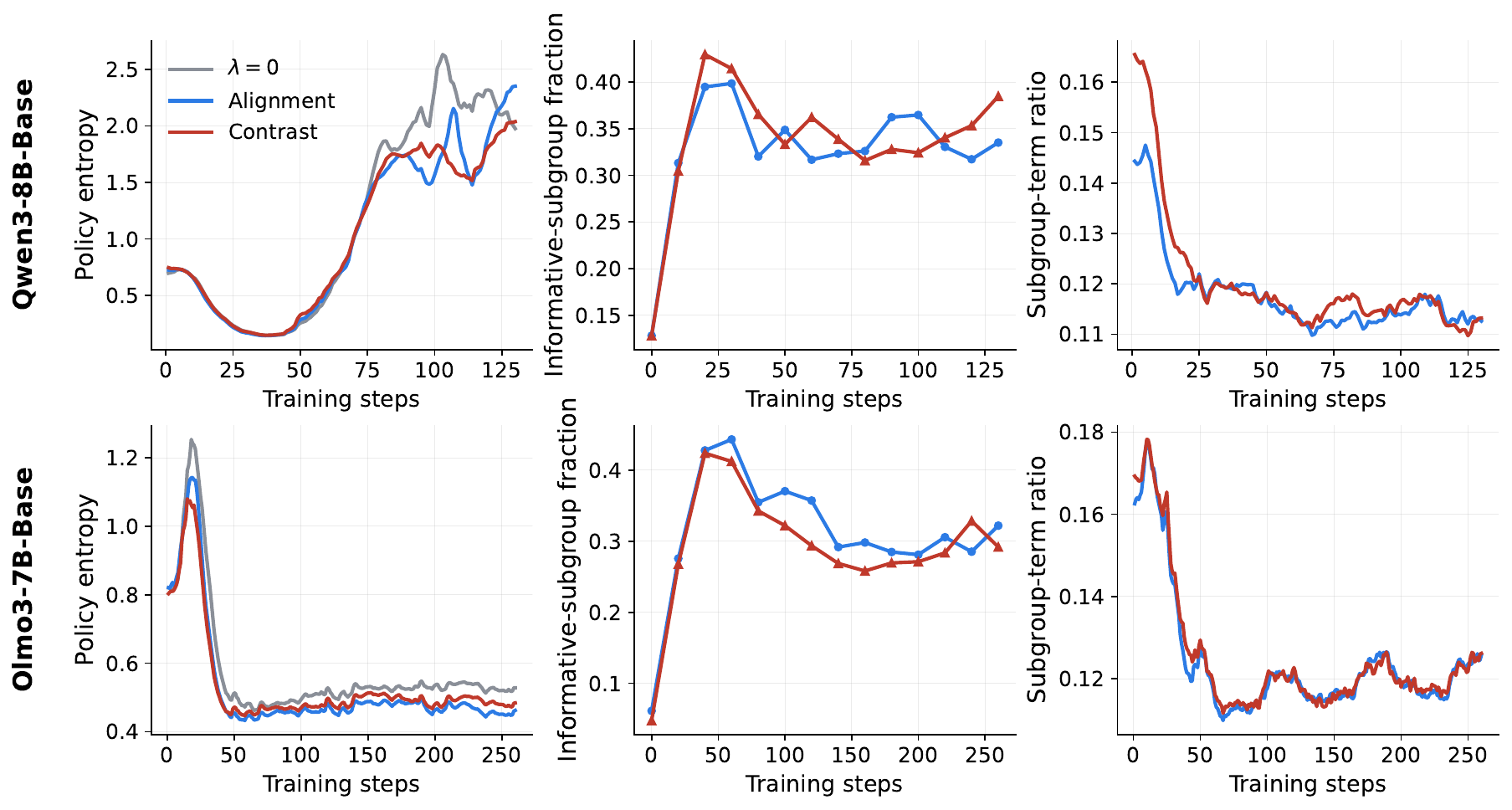}
\caption{Training dynamics on Qwen3-8B-Base (top) and Olmo3-7B-Base (bottom), for $\lambda{=}0$ (Strategy-GRPO), \textit{Alignment}, and \textit{Contrast}. \textbf{Left:} policy entropy over training (EMA-smoothed). \textbf{Middle:} fraction of prompts whose subgroups differ in success rate by at least $0.375$, i.e.\ by at least three of eight rollouts. \textbf{Right:} the magnitude of the subgroup term relative to the standard term in the advantage. The $\lambda{=}0$ baseline is included in the entropy panels for reference; the subgroup term is identically zero at $\lambda{=}0$, so the subgroup-term-ratio panels show only the two adjustment directions.}
\label{fig:dynamics}
\end{figure*}

\subsection{Ablation Study}

\paragraph{\textit{Alignment} and \textit{Contrast} respond differently to $\lambda$.} 
We ablate the coefficient $\lambda$ by scanning it over $\{-0.7,-0.5,-0.3,0,0.3,0.5,0.7\}$ on Qwen3-8B-Base, with results reported in Table~\ref{tab:lambda}. 
The sign of $\lambda$ selects the direction and its magnitude the strength; the two directions, however, respond to this strength in different ways. 

Under \textit{Alignment} ($\lambda>0$), the trend is non-monotonic: averaged over the four benchmarks, $\lambda{=}0.5$ is the weakest setting ($22.3$ pass@1, $47.9$ pass@16), while $\lambda{=}0.3$ and $\lambda{=}0.7$ are level on pass@1 (both $22.9$) and separate only on pass@16 ($51.2$ versus $49.4$). 
We speculate that the two extremes each lean on one dominant reference while keeping the other as a secondary correction: $\lambda{=}0.3$ stays anchored to the global mean and $\lambda{=}0.7$ to the subgroup, so credit follows a clear signal, whereas at $\lambda{=}0.5$ the two references carry equal weight and settling on neither leaves the midpoint weakest. 

Under Contrast ($\lambda<0$), the trend is instead monotonic: $\lambda{=}-0.3$ performs best, and both pass@1 and pass@16 fall steadily through $-0.5$ to $-0.7$. 
As $|\lambda|$ grows, the shared offset term increasingly dominates a rollout's own advantage, 
so credit is driven more by which subgroup it belongs to than by the rollout itself, blurring the distinctions the reward provides. 

In both directions, a moderate $|\lambda|$ works best. The coefficient governs how much subgroup structure enters the advantage, and an intermediate value supplements the per-rollout reward signal without letting the subgroup-level term overwhelm it.

\begin{table}[t]
\centering
\small
\setlength{\tabcolsep}{4.5pt}
\begin{tabular}{lcccc} 
\toprule 
Method & AIME24 & AIME25 & AIME26 & HMMT25 \\
\midrule 
Strategy-GRPO & 65.6 & 47.9 & 53.8 & 32.5 \\ 
\midrule
Alignment        & 66.4 & 49.8 & 56.0 & 32.7 \\ 
\quad + shuffle  & 64.5 $\downarrow$ & 47.2 $\downarrow$ & 51.7 $\downarrow$ & 29.2 $\downarrow$ \\ 
\midrule
Contrast         & 69.1 & 50.6 & 58.8 & 36.3 \\
\quad + shuffle  & 64.5 $\downarrow$ & 45.9 $\downarrow$  & 52.7 $\downarrow$  & 30.0 $\downarrow$ \\
\bottomrule
\end{tabular}
\caption{Offset-shuffling ablation on Qwen3-8B-Base (pass@16). Within each group, we permute the subgroup offsets so that each subgroup is assigned an offset from another, keeping the offset values unchanged. Arrows mark the change from the unshuffled setting.}
\label{tab:shuffle}
\end{table}            

\paragraph{Gains come from the correspondence between each subgroup and its offset.} 
To examine whether the gains arise from the correspondence between each subgroup and its offset, that is, whether the offset is a reliable proxy for subgroup quality, we conduct a shuffling ablation. 
For each subgroup, the offset is the deviation of its mean reward from the global mean; we retain the set of offsets within a group but permute their assignment across subgroups, so that each subgroup is referenced against an offset belonging to another. This keeps the magnitude and distribution of the offsets unchanged, altering only which subgroup each is attached to. 

As shown in Table~\ref{tab:shuffle}, once the assignment is permuted, both \textit{Alignment} and \textit{Contrast} degrade, even falling below the $\lambda{=}0$ baseline on all four benchmarks. 
This indicates that the gains depend on the pairing between each offset and its own subgroup, rather than on the mere presence of a nonzero offset term in the baseline. Since shuffling attaches actively wrong offsets, it shows that the offset is not an arbitrary perturbation, though it does not separate this from the milder possibility that any miscalibrated offset hurts.

\paragraph{Effect of strategy sampling.}
We ablate the strategy sampling to analyze how the sampling configuration affects the gains, varying two factors, as reported in Table~\ref{tab:sampling}.
The first is the plan sampling temperature ($1.0$ versus $0.7$); the second is how the $K$ strategies are selected, either by diversity or at random.
The default setting, temperature $1.0$ with diversity-based selection, attains the highest pass@16 on seven of the eight cells, and both departures reduce it, more so for \textit{Contrast}. 
The two departures are not ordered consistently with each other, which is unsurprising since they act differently: lowering the temperature narrows the pool of candidate strategies at the source, whereas random selection leaves the pool intact and only stops choosing dissimilar members from it. 
We therefore read these results as showing that the default configuration is the strongest of the three, rather than a monotone relationship between diversity and gain.

\begin{table}[t]
\centering 
\small
\setlength{\tabcolsep}{4.5pt}
\begin{tabular}{lcccc}
    \toprule
    Sampling & AIME24 & AIME25 & AIME26 & HMMT25 \\
    \midrule 
    \textit{Alignment} & & & & \\
    default  & 66.4 & 49.8 & 56.0 & 32.7 \\ 
    + $T{=}0.7$  & 67.9 $\uparrow$ & 48.6 $\downarrow$ & 53.3 $\downarrow$ & 31.3 $\downarrow$ \\ 
    + \textit{random}         & 66.3 $\downarrow$ & 48.1 $\downarrow$ & 54.0 $\downarrow$  & 29.4 $\downarrow$ \\ 
    \midrule
    \textit{Contrast} & & & & \\
    default & 69.1 & 50.6 & 58.8 & 36.3 \\
    + $T{=}0.7$              & 67.6 $\downarrow$ & 49.1 $\downarrow$ & 56.0 $\downarrow$ & 32.0 $\downarrow$ \\ 
    + \textit{random}  & 67.8 $\downarrow$ & 48.7 $\downarrow$ & 56.5 $\downarrow$ & 33.3 $\downarrow$ \\ 
    \bottomrule
\end{tabular}
\caption{Effect of strategy sampling on Qwen3-8B-Base (pass@16). Starting from the default setup (temperature $1.0$, diversity-based selection), we lower the sampling temperature to $0.7$ and, separately, replace diversity-based selection with random selection. Both changes weaken the diversity of the   
sampled strategies.}
\label{tab:sampling}
\end{table}

\subsection{Training Dynamics}
We examine the training dynamics of the two directions, tracking the policy entropy, the fraction of prompts with informative subgroups and the subgroup-term ratio over training in Figure~\ref{fig:dynamics}.

\paragraph{Policy entropy.} 
As shown in the left column of Figure~\ref{fig:dynamics}, the two models follow distinct entropy trajectories. 
On Qwen3-8B-Base, entropy first declines as the policy sharpens and then climbs through the rest of training, growing volatile in the later stage and trending toward entropy explosion; 
this onset is earliest under $\lambda{=}0$. 
On Olmo3-7B-Base, entropy instead rises to an early peak, falls back, and settles within a comparatively low and stable range. 
Across both models the $\lambda{=}0$ baseline sustains the highest entropy, with the two adjustment directions running at or below it. 
Overall, the adjustment reshapes credit without materially altering the exploration profile, consistent with its acting on the baseline alone rather than on the reward.

\paragraph{Informative subgroups. }
We next ask whether the subgroup structure the adjustment relies on persists over training. We periodically sample the plan-exec structure on a fixed set of prompts and call a prompt's subgroups \emph{informative} when their success rates span at least $0.375$, that is, three of eight rollouts. The middle column of Figure~\ref{fig:dynamics} tracks the fraction of informative prompts over training.  On both Qwen3-8B-Base and Olmo3-7B-Base, this fraction rises sharply once the model becomes competent, peaks in early training, and then relaxes to a stable level. It then holds steady rather than decaying, so the structure the adjustment acts on is a persistent feature of training rather than an artifact of the early phase. Where subgroup means coincide, $\delta_i{=}0$ and the baseline stays at $\mu_g$, so this curve tracks the share of prompts on which the adjustment is active, and on the rest it reduces to GRPO.

\paragraph{Contribution of the subgroup term.} 
Finally, we ask how much the subgroup offset contributes to the advantage over training.  For each rollout we take the ratio of the subgroup term $|\lambda\,\delta_i/\sigma_g|$ to the standard term $|A_i^{\mathrm{glob}}|$, averaged over rollouts,  as reported in the right column of Figure~\ref{fig:dynamics}. On both models the ratio is largest early in training and then settles into a low, stable range. The ratio is small for three reasons: $\lambda$ is moderate, the offsets $\delta_i$ are themselves modest, and the average runs over all rollouts, including those whose subgroups have near-equal  means and so contribute almost no subgroup term. Yet this term is far from negligible: it persists at a stable level throughout training, supplying subgroup information at every step without overwhelming the individual reward signal, especially on those informative groups.

\subsection{Discussion}

\paragraph{Alignment and Contrast.} 
Alignment and Contrast embody opposing philosophies of credit assignment. Alignment attenuates the between-subgroup component, drawing a rollout's credit toward its own subgroup, in the spirit of debiasing methods that normalize within homogeneous groups like Stratified-GRPO~\cite{zhu2025stratified}. 
Contrast does the opposite: it amplifies the between-subgroup component, directing credit toward rollouts from stronger subgroups, in the spirit of methods that reinforce more successful branches. 
A single signed coefficient unifies these opposing views as the two signs of one axis. 
Empirically, Contrast tends to yield the larger gains; 
we tentatively attribute this to the reward being a comparatively reliable indicator of strategy quality on verifiable mathematical tasks, so that emphasizing stronger subgroups is advantageous. 
We treat this as an empirical observation rather than an established mechanism. 

\paragraph{Generality.}
Although we induce subgroups from sampled strategies, the adjustment itself is agnostic to how the subgroups are formed.
Any partition of a group into subgroups defines a corresponding offset, and the same signed adjustment applies to subgroups formed in other ways, such as by clustering rollouts or by grouping on tool use or difficulty. 
What the algebra guarantees, however, is only that the adjustment is well defined for any partition; whether it helps depends on the partition yielding subgroups whose means are informative about quality, which we verified only for strategies on competition mathematics.

\paragraph{Limitations.} Our approach has two main limitations. First, it relies on explicitly constructed subgroups: we sample strategies before rollouts, which brings additional sampling cost, and the resulting subgroups may not match the distribution of direct sampling. 
We also inherit a domain assumption from this construction, since ``solution strategy'' is natural for competition mathematics but has no equally clear analogue in code generation, dialogue, or agentic tasks; our evaluation is confined to math reasoning and does not test the broader applicability suggested above.
Second, while a single signed coefficient unifies the two directions, we do not provide an automatic rule for choosing between them or for setting its magnitude;
the appropriate direction may depend on the task and the model. 
The sweep in Table~\ref{tab:lambda} was run only on Qwen3-8B-Base, so the conclusion that a moderate $|\lambda|$ works best is scoped to that model.
 We leave an adaptive choice of the coefficient, along with lighter-weight ways of obtaining subgroups, to future work.

%% file: sections/5_conclusion.tex
\section{Conclusion}
\label{sec:conclusion}

We revisited the group baseline in GRPO from the perspective of subgroups. 
When the rollouts of a prompt fall into heterogeneous subgroups of different reward distributions, the single mean-reward global baseline ignores this structure and assigns credit without regard to a rollout's standing within its subgroup or the standing of that subgroup among the rest. 
We introduced the \textit{subgroup offset}, the deviation of a subgroup's mean reward from the global mean, and adjusted the baseline along it by a single signed coefficient $\lambda$. 
Its sign yields two opposing directions: \textit{Alignment} shifts a rollout's credit toward its subgroup, while \textit{Contrast} accentuates the differences between subgroups. Both act on the baseline alone, leaving the reward distribution unchanged. 
On Qwen3-8B-Base and Olmo3-7B-Base across four competition-level math benchmarks, both directions improve modestly but consistently over the $\lambda{=}0$ anchor, which shares the same strategy-conditioned sampling, with \textit{Contrast} the stronger of the two and the gains concentrated in multi-sample coverage; ablations confirm that they depend on the genuine correspondence between subgroups and their offsets.
These results suggest that a subgroup-aware baseline offers a simple lever for credit assignment in RLVR.
We leave to future work an adaptive choice of both the direction and the strength of the adjustment, together with a broader study across domains and subgroup constructions.